\RequirePackage{iftex}

\ifLuaTeX

\fi

\documentclass[runningheads]{llncs}

\usepackage{eccv} 

\usepackage{eccvabbrv}
\usepackage{graphicx}
\usepackage{booktabs}
\usepackage[accsupp]{axessibility} 

\usepackage[pagebackref,breaklinks,colorlinks,citecolor=eccvblue]{hyperref} 
\usepackage{hyperref}
\usepackage{orcidlink}
\hypersetup{colorlinks=true,linkcolor=[rgb]{0.1,0.3,0.9},urlcolor=[rgb]{0.2,0.2,0.2},citecolor=[rgb]{0.1,0.3,0.9}}

\usepackage{orcidlink}
\usepackage{bbm}

\begin{document}

\title{Do 3D Medical Foundation Models See Through MRI Artifacts? \\ A Controlled Study of Representation Robustness} 

\titlerunning{3D Foundation Models Under MRI Artifacts}

\author{Julia Anna Mielcarz \orcidlink{0009-0005-3992-4774} \and
Daniel Klaaby \and
Mostafa Mehdipour Ghazi \orcidlink{0000-0002-8473-281X}}

\authorrunning{J. A. Mielcarz et al.}

\institute{Pioneer Centre for AI, University of Copenhagen, Copenhagen, Denmark \\
\email{ghazi@di.ku.dk}}

\maketitle

\begin{abstract} 

Self-supervised 3D medical foundation models are increasingly used as general-purpose feature extractors, yet their sensitivity to MRI artifacts remains poorly understood. We present a controlled evaluation of representation robustness across five pretrained 3D encoders spanning different architectures, objectives, pretraining domains, and dataset scales. Using BraTS-Africa cases with four MRI sequences, we generate seven frequency- and image-domain artifacts at five predefined corruption settings. Robustness is assessed using linear centered kernel alignment (CKA), RankMe, and UMAP, complemented by an independent segmentation-consistency analysis. We find that robustness is strongly model- and artifact-dependent. 3DINO exhibits the most consistently stable representations, while BrainIAC is highly sensitive to several corruptions; NeuroVFM, BrainFM, and Neuro-SimCLR show intermediate but distinct artifact-specific profiles. Across many conditions, CKA decreases substantially while RankMe remains comparatively stable, indicating that artifacts often distort representation geometry without causing dimensional collapse. Segmentation consistency also degrades under corruption, particularly for ghosting and Rician noise, but aligns only partially with representation-level robustness. These findings show that larger-scale or domain-specific pretraining alone does not guarantee artifact invariance and motivate explicit robustness evaluation before deploying 3D foundation models in heterogeneous MRI settings.


\end{abstract}

\section{Introduction} \label{sec:intro}

Self-supervised pretraining has enabled 3D medical-image encoders to learn reusable representations from large collections of unlabeled scans. Recent models span broad multi-organ pretraining \cite{xu2025generalizable}, brain-specific contrastive learning \cite{tak2026generalizable,kaczmarek2025building}, multi-task learning \cite{liu2025modality}, and large-scale joint-embedding prediction \cite{kondepudi2026health}. These representations are increasingly used as general-purpose features for segmentation, classification, and disease analysis. Their practical value, however, depends not only on downstream accuracy but also on whether they remain stable under variations that are unrelated to the anatomy or pathology of interest.

This requirement is particularly important in MRI, where scanner hardware, acquisition protocols, reconstruction procedures, and patient motion can substantially alter image appearance. Common degradations include noise, intensity inhomogeneity, ghosting, Gibbs ringing, blurring, and k-space interference \cite{krupa2015artifacts}. Motion-related degradation alone affects a substantial fraction of clinical MR acquisitions and can require repeated scanning \cite{andre2015toward}. Such artifacts may provide high-amplitude visual cues that pretrained encoders inadvertently represent, despite carrying no intended clinical information. This creates a risk that feature spaces change across acquisition conditions even when the underlying anatomy is unchanged, limiting transfer across scanners, sites, and patient populations \cite{geirhos2020shortcut,glocker2019machine}.

Robustness to these effects cannot be inferred reliably from a single downstream task. Task performance depends on the prediction head, available annotations, and task-specific information, and may obscure substantial changes within the underlying representation. Direct representation analysis instead asks whether matched clean and corrupted images retain similar relational geometry and spectral structure. Although representation similarity has been extensively studied in general deep learning, e.g., using Representational Similarity Analysis (RSA) \cite{kriegeskorte2008representational} or Singular Vector Canonical Correlation Analysis (SVCCA) \cite{raghu2017svcca}, the sensitivity of recent 3D medical and neuroimaging encoders to controlled MRI artifacts remains poorly understood. In particular, it is unclear whether robustness is associated with broader pretraining data, greater training scale, brain-specific specialization, architectural choice, or pretraining objective.

We address this gap through a controlled benchmark of five public 3D encoders with different architectures, objectives, domains, and pretraining scales. Clean brain MRIs are perturbed using seven frequency- and image-domain artifacts at five predefined corruption settings and evaluated across four modalities. We quantify changes in representation geometry using centered kernel alignment (CKA) \cite{kornblith2019similarity}, changes in spectral effective rank using RankMe \cite{garrido2023rankme}, and qualitative organization using UMAP \cite{mcinnes2018umap}. An independent segmentation-consistency experiment provides a task-level reference for assessing whether the relative effects of the same artifacts extend beyond the evaluated representation spaces.

Our main \textbf{contributions} are three-fold: (1) We introduce a controlled evaluation of artifact robustness across five recent pretrained 3D medical-image encoders, seven MRI-relevant corruptions, five corruption settings, and four MRI modalities (T1, T1ce, T2, FLAIR). (2) We show that robustness is strongly model- and artifact-dependent: broad medical pretraining provides the most consistent stability overall, while brain-specific specialization or larger pretraining scale does not guarantee robustness to all corruption mechanisms. (3) We find that artifacts often induce substantial geometric drift without a corresponding collapse in effective rank \cite{del2021effective}, demonstrating that representation similarity and spectral dimensionality capture distinct failure modes. Comparison with an independent segmentation model further shows that representation-level and task-level sensitivities align only partially.

\section{Related Work} \label{sec:related}

\paragraph{\textbf{Pretrained 3D medical-image encoders.}}
Early medical-image transfer learning largely relied on two-dimensional networks pretrained on natural images, despite the different visual statistics and volumetric structure of medical data. Recent work instead pretrains 3D encoders directly on medical images. 3DINO uses self-distillation across approximately 100,000 multi-organ MRI, CT, and PET volumes; BrainIAC and Neuro-SimCLR apply contrastive learning to brain MRI; BrainFM learns a shared encoder through multiple brain-imaging tasks; and NeuroVFM uses volumetric joint-embedding prediction at substantially larger clinical scale. These models differ simultaneously in architecture, objective, pretraining domain, and dataset scale, but their relative sensitivity to MRI artifacts has not been systematically compared.

\paragraph{\textbf{MRI artifacts and robustness.}}
MRI quality can be degraded by acquisition and reconstruction effects such as noise, intensity inhomogeneity, radiofrequency interference, motion ghosting, and finite k-space sampling. Prior studies have simulated such effects for data augmentation, quality assessment, artifact detection, or robustness evaluation. Most evaluations, however, quantify image quality or downstream predictive performance. These measurements do not directly reveal whether matched clean and corrupted images retain a similar organization within the representation space of a pretrained encoder. Our study instead uses controlled artifacts to compare representation robustness across multiple recent 3D medical-image models.

\paragraph{\textbf{Representation-level evaluation.}}
Neural representations can be evaluated through their similarity, geometry, or spectral structure. RSA, SVCCA, and CKA compare feature spaces derived from matched inputs. CKA is particularly suitable as it measures alignment between centered Gram structures while remaining invariant to orthogonal transformations and isotropic scaling. Complementarily, RankMe estimates spectral effective rank from the entropy of normalized singular values, providing a label-free measure of whether embeddings become concentrated in fewer directions. We use them jointly, as an artifact may distort the relative geometry of the subjects without any dimensional collapse. UMAP is used only for qualitative inspection of artifact-dependent organization.

\section{Methods} \label{sec:methods}
 
We investigate how MRI artifacts alter the internal representations of pretrained 3D medical-image encoders and whether this sensitivity varies across pretraining strategies, artifact types, and corruption settings. To this end, we construct a controlled-perturbation benchmark and evaluate five frozen encoders using complementary measures of representational geometry and spectral structure. An independent tumor-segmentation consistency analysis provides a task-level reference for comparing the relative effects of the same artifacts.

\subsection{Pretrained 3D Encoders} \label{sec:models}

\paragraph{\textbf{Model selection.}}
We evaluate five publicly available pretrained models that process three-dimensional medical images. The selected models differ along four design dimensions that may influence robustness to imaging artifacts: (i) network architecture, including vision transformers (ViTs), convolutional encoders, and U-Net-like dense-prediction networks; (ii) pretraining objective, including self-distillation, instance-contrastive learning, joint-embedding predictive learning, and multi-task learning; (iii) pretraining domain, ranging from multi-organ, multimodal medical imaging to brain-specific MRI; and (iv) pretraining scale, ranging from tens of thousands to several million images. The purpose of this comparison is not to rank the models by downstream accuracy, but to determine whether these design choices are associated with different degrees of representational stability under controlled image corruptions.

All pretrained models are frozen throughout the study. For each model $m$, an image volume $\mathcal{I}$ is transformed by the model-specific preprocessing operator $\mathcal{P}_m$ and then passed through the frozen encoder $f_m$. A fixed aggregation operator $g_m$ converts the token sequence or spatial feature map into a single vector: \begin{equation}
\mathbf{z}^{(m)} = \phi_m(\mathcal{I}) = g_m\!\left(f_m\!\left(\mathcal{P}_m(\mathcal{I})\right) \right) \in \mathbb{R}^{d_m}, 
\label{eq:model_embedding} 
\end{equation}
where $d_m$ denotes the embedding dimensionality. For each model, identical preprocessing and feature extraction are applied to the clean and corrupted volumes.

We evaluate five complementary pretrained 3D encoders. \textbf{3DINO} \cite{xu2025generalizable} extends DINO-style self-distillation \cite{oquab2024dinov2} to volumetric, multi-organ medical imaging using a ViT backbone. \textbf{BrainIAC} \cite{tak2026generalizable} is a brain-MRI-specific ViT pretrained with the SimCLR objective \cite{chen2020simple}. \textbf{BrainFM} \cite{liu2025modality} adopts a U-Net-like architecture trained jointly across multiple brain-imaging tasks. \textbf{NeuroVFM} \cite{kondepudi2026health} is a large-scale ViT-based neuroimaging encoder pretrained on heterogeneous clinical MRI and CT volumes using Volumetric Joint-Embedding Predictive Architecture (Vol-JEPA). Finally, \textbf{Neuro-SimCLR} \cite{kaczmarek2025building} uses a convolutional 3D ResNet-18 pretrained with SimCLR on structural brain MRI. Together, these models span different architectures, pretraining objectives, domains, and dataset scales.

\Cref{tab:models} summarizes the properties of the original studies and the representations extracted in this work. Because the source studies use different definitions of a training sample, we retain their original terminology, such as scans, volumes, or acquired images, rather than interpreting these counts as numbers of unique subjects. Full details on the pretraining data, model-specific preprocessing, and feature-extraction layers are provided in \textbf{Appendix}.

\begin{table}[t] 
\caption{Pretrained 3D encoders evaluated in this study. ``Representation'' denotes the frozen feature extracted in our experiments, with dimensionality $d$, and is not necessarily the representation used in the original downstream evaluations.} 
\vspace{-0.2cm}
\label{tab:models}
\centering
\renewcommand{\arraystretch}{1.2}
\resizebox{\textwidth}{!}{
\begin{tabular}{lccccc} 
\toprule
Model & Backbone & Objective & Domain & Pretraining data & Representation ($d$) \\ 
\midrule 
3DINO \cite{xu2025generalizable} & 3D ViT-L/16 & Self-distillation & Multi-organ MRI/CT/PET & 100,000 volumes & Final class token (1024) \\ 
BrainIAC \cite{tak2026generalizable} & 3D ViT-B/16 & Contrastive & Brain MRI & 32,015 scans & Final class token (768) \\
BrainFM \cite{liu2025modality} & 3D U-Net-like & Multi-task learning & Brain MRI/CT & 7,800 real images & Globally pooled encoder features (2048) \\
NeuroVFM \cite{kondepudi2026health} & 3D ViT-B/$(4{\times}16{\times}16)$ & Volumetric JEPA & Neuroimaging MRI/CT & 5.24M volumes & Mean-pooled patch tokens (768) \\
Neuro-SimCLR \cite{kaczmarek2025building} & 3D ResNet-18 & Contrastive & Brain MRI & 44,958 scans & Globally pooled encoder features (512) \\
\bottomrule \end{tabular}
} 
\end{table}

\subsection{Artifact Simulation Framework} \label{sec:artifacts}

We construct controlled corrupted versions of each clean MRI volume to evaluate how pretrained representations respond to image-quality degradation. The selected transforms are divided according to the domain in which they are implemented: (i) \emph{frequency-domain corruptions}, which modify the Fourier representation of the image, and (ii) \emph{image-domain corruptions}, which operate directly on voxel intensities. The former comprises k-space spikes, periodic k-space ghosting, and k-space truncation leading to Gibbs ringing. The latter comprises Rician noise, Gaussian blur, bias field, and gamma contrast transformation.

Let $\mathcal{I}\in[0,1]^{H\times W\times D}$ denote a real-valued magnitude MRI volume after common intensity normalization, but before model-specific preprocessing. Each artifact transformation is applied to this common representation, after which the resulting volume is processed independently by the preprocessing operator $\mathcal{P}_m$ of each encoder. Thus, all models are evaluated using corruptions generated from the same underlying image.

For artifact type $a$ and severity level $\ell$, the corrupted image is written as
\begin{equation}
\mathcal{I}^{(a,\ell)} = \mathcal{T}_{a} \left(\mathcal{I}; \theta_{a,\ell}, \boldsymbol{\xi}_{a}\right),
\label{eq:artifact_general}
\end{equation}
where $\theta_{a,\ell}$ is the parameter controlling artifact severity and $\boldsymbol{\xi}_{a}$ contains any random variables, such as spike locations or noise realizations. Each artifact is applied independently so that changes in the resulting representations can be attributed to one corruption type and severity level.

\subsubsection{Frequency-Domain Corruptions} \label{sec:frequency_artifacts}

Let $\mathcal{F}$ and $\mathcal{F}^{-1}$ denote the three-dimensional fast Fourier transform (FFT) and its inverse, respectively. We use the centered Fourier representation
\begin{equation}
\mathcal{K} = \operatorname{fftshift} \left(\mathcal{F}(\mathcal{I})\right) \in\mathbb{C}^{H\times W\times D},
\label{eq:centered_kspace}
\end{equation}
where the zero-frequency component is located at the center of the array. Given a modified spectrum $\widetilde{\mathcal{K}}$, the corresponding magnitude image is reconstructed as
\begin{equation}
\mathcal{R}(\widetilde{\mathcal{K}}) = \left| \mathcal{F}^{-1} \left(\operatorname{ifftshift}(\widetilde{\mathcal{K}})\right)\right|.
\label{eq:kspace_reconstruction}
\end{equation}

\paragraph{\textbf{K-space spikes.}}
Transient radiofrequency interference or hardware instability can introduce isolated, high-amplitude samples into k-space. Because a single Fourier coefficient contributes a sinusoidal component over the full spatial domain, such disturbances appear as global stripe or herringbone patterns in the reconstructed image \cite{krupa2015artifacts}. We sample a set of $M$ distinct locations
$\Omega_{\mathrm{sp}} = \{ \mathbf{k}_1, \ldots, \mathbf{k}_M\}$ from the k-space grid. The corrupted image is defined as
\begin{equation}
\mathcal{K}_{\mathrm{sp}}[\mathbf{k}] = \mathcal{K}[\mathbf{k}] + \alpha_{\mathrm{sp}} \Vert\mathcal{K}\Vert_{\infty} \mathbbm{1} \left[\mathbf{k}\in\Omega_{\mathrm{sp}}\right], \quad
\mathcal{T}_{\mathrm{sp}}(\mathcal{I}) = \mathcal{R}(\mathcal{K}_{\mathrm{sp}})
\label{eq:spike_kspace}
\end{equation}
where $\Vert\mathcal{K}\Vert_{\infty} = \max_{\mathbf{k}}|\mathcal{K}[\mathbf{k}]|$,
$\alpha_{\mathrm{sp}}>0$ controls the spike amplitude, and $\mathbbm{1}[\cdot]$ is the indicator function. We fix the number of injected coefficients to $M=2$ and vary only $\alpha_{\mathrm{sp}}$ across severity levels, thereby separating artifact amplitude from the number of corrupted k-space locations.

\paragraph{\textbf{Periodic k-space ghosting.}}
Ghost replicas can arise when periodically inconsistent measurements occur along the phase-encoding direction \cite{wood1985mr,krupa2015artifacts}. A common synthetic approximation is to periodically attenuate k-space lines, producing displaced copies of the image after inverse Fourier reconstruction. Let $q\in\{x,y,z\}$ denote the selected encoding axis, $k_q$ be the corresponding frequency index, and $R$ denote the modulation period. The periodic modulation is obtained as
\begin{equation}
\mathcal{K}_{\mathrm{gh}}[\mathbf{k}] =
\begin{cases}
\alpha \mathcal{K}[\mathbf{k}], & k_q \equiv 0 \pmod R, \\
\mathcal{K}[\mathbf{k}], & \text{otherwise},
\end{cases}
\quad
\mathcal{T}_{\mathrm{gh}}(\mathcal{I}) = \mathcal{R}(\mathcal{K}_{\mathrm{gh}}).
\label{eq:ghost_mask}
\end{equation}
We fix $R=3$, set $q$ to $x$-axis, and vary $\alpha_{\mathrm{gh}}\in(0,1]$. Smaller values cause stronger attenuation of the periodically selected lines and consequently more prominent replicas. This transformation is referred to as periodic k-space ghosting rather than a complete physical simulation of subject motion, because it does not model a time-dependent rigid transformation during acquisition.

\paragraph{\textbf{Gibbs ringing.}}
Gibbs ringing results from the finite sampling or truncation of k-space. Multiplication of k-space by a rectangular window corresponds to convolution of the reconstructed image with a sinc-like point-spread function, producing oscillations near sharp intensity transitions \cite{block2008suppression,kellner2016gibbs}. Following previous MRI augmentation frameworks \cite{mehdipour2025fast,van2024non}, we simulate this effect by retaining only a centered region of k-space. Let $c_q$ denote the number of retained coefficients along the axis $q$. The truncation mask is
\begin{equation}
\mathcal{K}_{\mathrm{gi}}[\mathbf{k}] = \prod_{q\in\{x,y,z\}} \mathbbm{1} \left[\left| k_q-\frac{N_q-1}{2} \right| \leq \frac{c_q-1}{2} \right] \cdot \mathcal{K}[\mathbf{k}], \quad
\mathcal{T}_{\mathrm{gi}}(\mathcal{I}) = \mathcal{R}(\mathcal{K}_{\mathrm{gi}})
\label{eq:gibbs_mask}
\end{equation}
where $(N_x,N_y,N_z)=(H,W,D)$. We use the retained width along the $y$-axis. Smaller cutoff values of $c$ remove a larger fraction of peripheral high-frequency coefficients and therefore increase both ringing and loss of fine spatial detail. These two effects are inherent to rectangular frequency truncation and are not treated as separate corruptions.

\subsubsection{Image-Domain Corruptions} \label{sec:image_artifacts}

\paragraph{\textbf{Bias field.}}
MRI intensity inhomogeneity is modeled as a smooth multiplicative field applied to the underlying image \cite{sled1998nonparametric,perez2021torchio}. Let $(u_x,u_y,u_z)$ denote voxel-index coordinates on the lattice $\Omega = \{1,\ldots,H\} \times \{1,\ldots,W\} \times \{1,\ldots,D\}$, and let $(x_0,y_0,z_0)$ denote the center of the field. Following previous MRI augmentation studies that employ an elliptic gradient field \cite{hui2010fast,mehdipour2025fast,van2024non}, we define a paraboloidal spatial field and apply it to the image to obtain the corrupted one as
\begin{equation}
B(\mathbf{u}) = 1 - \lambda \left[\frac{(u_x-x_0)^2}{H^2} + \frac{(u_y-y_0)^2}{W^2} + \frac{(u_z-z_0)^2}{D^2}\right],
\quad
\mathcal{T}_{\mathrm{bf}}(\mathcal{I}) = B\odot\mathcal{I},
\label{eq:bias_field}
\end{equation} 
where $\odot$ denotes element-wise multiplication and $\lambda\geq0$ controls the strength of the intensity inhomogeneity. The gain equals 1 at the field center and, by fixing the paraboloid center at the volume midpoint, can decrease smoothly to a minimum value of approximately $1-0.75\lambda$ at the most distant (corner) voxels.

\paragraph{\textbf{Rician noise.}}
In conventional magnitude MRI, thermal noise can be approximated as independent zero-mean Gaussian noise in the real and imaginary components of the complex signal. Taking the magnitude of the resulting complex image leads to Rician-distributed voxel intensities. Assuming a zero-phase noiseless signal, we simulate this process as
\begin{equation}
\mathcal{T}_{\mathrm{ri}}(\mathcal{I}) = \sqrt{\left( \mathcal{I} + \sigma_{\mathrm{ri}}\boldsymbol{\epsilon}_{1} \right)^2 + \left( \sigma_{\mathrm{ri}}\boldsymbol{\epsilon}_{2} \right)^2}, \quad
\boldsymbol{\epsilon}_{1}, \boldsymbol{\epsilon}_{2}
\overset{\mathrm{i.i.d.}}{\sim} \mathcal{N}(\mathbf{0},\mathbf{I}),
\label{eq:rician}
\end{equation}
where $\sigma_{\mathrm{ri}}$ is expressed relative to the normalized intensity range and controls noise severity. Unlike additive Gaussian noise, this construction introduces a positive noise floor in low signal regions.

\paragraph{\textbf{Gaussian blur.}}
We model smooth degradation of apparent spatial resolution by convolving the volume with a normalized isotropic 3D Gaussian kernel:
\begin{equation}
\mathcal{G}_{\sigma_{bl}}(\mathbf{u}) = \frac{1}{(2\pi\sigma^2_{bl})^{3/2}} \exp \left(- \frac{\Vert\mathbf{u}\Vert_2^2}{2\sigma^2_{bl}} \right), \quad
\mathcal{T}_{\mathrm{bl}}(\mathcal{I}) = \mathcal{G}_{\sigma_{\mathrm{bl}}} * \mathcal{I},
\label{eq:gaussian_blur}
\end{equation}
where $*$ denotes convolution and the parameter $\sigma_{\mathrm{bl}}$, measured in voxels, determines the degree of smoothing. Since the Fourier transform of a Gaussian is also Gaussian, increasing $\sigma_{\mathrm{bl}}$ suppresses high-spatial-frequency image content.

\paragraph{\textbf{Gamma contrast.}}
We simulate global nonlinear changes in image contrast using a power-law transformation:
\begin{equation}
\mathcal{T}_{\gamma}(\mathcal{I}) = \mathcal{I}^{\gamma},
\label{eq:gamma}
\end{equation}
where the exponentiation is applied voxel-wise. For intensities normalized to $[0,1]$, $\gamma<1$ increases intermediate intensities, whereas $\gamma>1$ suppresses them. Gamma correction is used here as a controlled image-domain contrast perturbation and not as a physical forward model of an MRI pulse-sequence parameter.

\paragraph{\textbf{Output range.}}
Some transformations, such as Rician noise and k-space spikes, may produce values outside the initial unit interval. To maintain a valid input
range for subsequent processing, the transformed image is clipped voxel-wise:
\begin{equation}
\widehat{\mathcal{T}}_{a}(\mathcal{I}) = \min\!\bigl(\max(\mathcal{T}_a(\mathcal{I}),0),1\bigr).
\label{eq:artifact_clipping}
\end{equation}
The same operation is applied at all severity levels before model preprocessing.

\paragraph{\textbf{Severity levels.}}
Each corruption is evaluated at five predefined severity levels, denoted L1--L5. The parameter values are summarized in \Cref{tab:severity}. They were selected through a preliminary visual inspection to cover a broad range, from subtle degradation to pronounced corruption, while retaining recognizable anatomical structure. The levels should therefore be interpreted as controlled experimental perturbations rather than calibrated clinical image-quality grades.

For Rician noise, Gaussian blur, k-space spikes, and the multiplicative bias field, larger parameter values produce stronger perturbations. For periodic k-space ghosting, smaller values of $\alpha_{\mathrm{gh}}$ produce stronger attenuation of the periodically selected k-space lines. Similarly, for Gibbs ringing, smaller values of the retained width $c$ remove a larger portion of peripheral k-space and therefore produce stronger truncation effects. Finally, gamma includes values both below and above the identity transform $\gamma=1$. They represent different nonlinear contrast conditions rather than a single unidirectional severity sequence: values below one increase intermediate intensities, whereas values above one suppress them.

\begin{table}[t]
\caption{Parameters used for the five predefined corruption settings. For ghosting and Gibbs ringing, stronger corruption corresponds to smaller parameter values. The gamma settings span both sides of the identity transformation $\gamma=1$ and therefore represent contrast conditions rather than a monotonic severity sequence.}
\vspace{-0.2cm}
\label{tab:severity}
\centering
\renewcommand{\arraystretch}{1.1}
\begin{tabular}{lccccc}
\toprule
Artifact (parameter) & L1 & L2 & L3 & L4 & L5 \\
\midrule
Rician noise ($\sigma_{\mathrm{ri}}$) & 0.07 & 0.08 & 0.09 & 0.10 & 0.15 \\
Gaussian blur ($\sigma_{\mathrm{bl}}$) & 0.60 & 0.80 & 1.00 & 1.20 & 1.40 \\
Bias field strength ($\lambda$) & 0.60 & 0.80 & 1.00 & 1.20 & 1.40 \\
Gamma contrast ($\gamma$) & 0.40 & 0.60 & 0.80 & 1.75 & 2.00 \\
Spike amplitude ($\alpha_{\mathrm{sp}}$) & 0.45 & 0.60 & 0.75 & 0.90 & 1.05 \\
Ghosting attenuation ($\alpha_{\mathrm{gh}}$) \qquad & 0.70 & 0.60 & 0.55 & 0.50 & 0.45 \\
Gibbs cutoff width ($c$) & 169 & 153 & 138 & 122 & 107 \\
\bottomrule
\end{tabular}
\end{table}

\subsection{Representation-Level Robustness Metrics} \label{sec:metrics}

We evaluate artifact robustness in the representation space of each frozen encoder. Downstream performance cannot distinguish whether a corruption alters the pretrained representation or whether the observed performance change arises from the task-specific prediction model. Representation-level analysis enables label-free comparison of the same images before and after corruption. We characterize two complementary properties: preservation of the relational geometry of the embedding space using linear centered kernel alignment (CKA), and changes in its singular-value spectrum using RankMe. We then use Uniform Manifold Approximation and Projection (UMAP) for qualitative visualization.

For a given encoder $m$, let $\mathbf{z}^{(m)}_n = \phi_m(\mathcal{I}_n) \in \mathbb{R}^{d_m}$ denote the embedding of clean image $\mathcal{I}_n$, and let $\mathbf{z}^{(m,a,\ell)}_n = \phi_m\!\left( \mathcal{T}_{a,\ell}(\mathcal{I}_n) \right) \in \mathbb{R}^{d_m}$ denote the embedding of the corresponding image corrupted by artifact $a$ at setting $\ell$. For $N$ matched images, the clean and corrupted embeddings are stacked row-wise as
\begin{equation} 
Z^{(m)} = 
\begin{bmatrix} 
\mathbf{z}^{(m)}_1, \dots, \mathbf{z}^{(m)}_N 
\end{bmatrix}^\top, \qquad
Z^{(m,a,\ell)} = 
\begin{bmatrix} 
\mathbf{z}^{(m,a,\ell)}_1, \dots, \mathbf{z}^{(m,a,\ell)}_N 
\end{bmatrix}^\top. 
\label{eq:embedding_matrices} 
\end{equation}

\paragraph{\textbf{Representational similarity.}} 
Linear CKA compares two representations via their centered pairwise inner-product structures \cite{kornblith2019similarity}. It is invariant to isotropic scaling and orthogonal transformations of the feature coordinates, making it suitable for measuring whether the relative organization of samples is preserved despite changes in the feature basis.

Let $C_N = I_N-\frac{1}{N}\mathbf{1}\mathbf{1}^{\top}$ denote the centering matrix, and define $X=C_N Z^{(m)}$ and $Y=C_N Z^{(m,a,\ell)}$. Linear CKA is computed as \begin{equation} 
\operatorname{CKA}(X,Y) = \frac{ \left\|X^\top Y\right\|_F^2 }{ \left\|X^\top X\right\|_F \left\|Y^\top Y\right\|_F }, 
\label{eq:cka} 
\end{equation} 
where $\|\cdot\|_F$ denotes the Frobenius norm, and the numerator measures the alignment between the centered linear Gram matrices of the two representations.

CKA lies in $[0,1]$ for the positive-semidefinite linear kernels here. A value close to one indicates that the pairwise geometry of the clean embedding set is preserved after corruption, up to the invariances of linear CKA. Lower values indicate increasing distortion of the relational structure among images. We compute CKA separately for every encoder, MRI modality, artifact type, and corruption setting, using matched clean and corrupted images in the same order.

\paragraph{\textbf{Spectral effective rank.}}
CKA measures correspondence between clean and corrupted representations but does not describe how their variance is distributed across feature directions. A corruption may concentrate the embeddings into fewer dominant directions without destroying their pairwise organization. We therefore compute RankMe, an entropy-based effective rank derived from the singular-value spectrum of the representation matrix \cite{garrido2023rankme}.

For a zero-centered embedding matrix $Z\in\mathbb{R}^{N\times d_m}$, let $Z = U\Sigma V^\top$, with singular values $\Sigma = \operatorname{diag}(\sigma_1,\ldots,\sigma_r)$, where $r\leq\min(N,d_m)$ and $\sigma_j\geq0$. The normalized singular values and RankMe scores are then calculated as
\begin{equation} 
p_j = \frac{\sigma_j} {\sum_{k=1}^{r}\sigma_k} + \varepsilon, \quad
\operatorname{RankMe}(Z) = \exp\left( -\sum_{j=1}^{r}p_j\log p_j \right),
\label{eq:rankme} 
\end{equation} 
where $\varepsilon$ is a small constant used to avoid evaluating $\log 0$. RankMe is defined as

RankMe is largest when the singular-value mass is distributed uniformly across the available directions and decreases as the spectrum becomes concentrated in a smaller number of dominant directions. We therefore use it to quantify corruption-associated spectral concentration or expansion of each model's embedding space. A decrease relative to the clean representation indicates that fewer directions dominate the corrupted embeddings; it should not, however, be interpreted directly as loss of semantic information. 

Because the maximum attainable effective rank depends on both the number of images and the feature dimensionality, RankMe is interpreted primarily by comparing each model with its own baseline across artifact settings, rather than by treating absolute RankMe values as directly comparable measures of representation quality across different architectures.

\paragraph{\textbf{Qualitative visualization.}}
To visualize corruption-associated changes that may not be fully summarized by CKA or RankMe, we use UMAP \cite{mcinnes2018umap}. It constructs a neighborhood graph in the original feature space and optimizes a low-dimensional representation intended to preserve its local neighborhood structure. 

For each encoder and MRI modality, clean and corrupted embeddings are included in a common UMAP fit so that their two-dimensional coordinates share the same projection. To reduce visual overlap, the qualitative analysis includes the clean images and corruption settings L3--L5. Embeddings generated by the same artifact type are assigned a common visual category, allowing inspection of whether artifact-corrupted images remain close to their clean counterparts or form artifact-specific regions in the projected space. UMAP is run with the default setting and is used only as an exploratory visualization. 

\subsection{Segmentation Robustness Analysis} \label{sec:downstream}

The representation-level metrics quantify changes within each pretrained encoder but do not establish whether the same image corruptions also affect a clinically relevant prediction task. We therefore conduct a complementary segmentation experiment using TumorSynth \cite{wu2026tumorsynth}, an independently developed model for joint segmentation of brain anatomy and tumors. In whole-tumor mode, TumorSynth produces 18 tissue classes (labels 1--17 for healthy tissue and label 18 for whole tumor). It does not use the representations extracted from any of the five encoders. Consequently, this experiment is not a downstream evaluation of those foundation models. Instead, it provides an independent task-level reference for examining whether artifacts that substantially perturb pretrained representations also tend to disrupt an automated image-analysis pipeline.

Let $S(\mathcal{I}_n) = \widehat{Y}_n$ denote the whole-label mask predicted by TumorSynth from clean image $\mathcal{I}_n$ and $S\!\left( \mathcal{T}_{a,\ell}(\mathcal{I}_n) \right) = \widehat{Y}^{(a,\ell)}_n$ be the prediction obtained using the same TumorSynth model after applying artifact $a$ at setting $\ell$.

\paragraph{\textbf{Segmentation consistency.}}
Because the simulated artifacts modify image intensities without changing the spatial coordinate system or anatomical geometry, we evaluate segmentation consistency using the Dice similarity coefficient between $\widehat{Y}^{(a,\ell)}_n$ and $\widehat{Y}_n$. A value close to one indicates that TumorSynth produces similar masks from the clean and corrupted images. This measure is interpreted only as prediction stability under corruption and not as segmentation accuracy. 

%

\subsection{Evaluation Data} \label{sec:data}

We use 95 diffuse-glioma cases from the BraTS-Africa dataset \cite{adewole2025brats}, a retrospective multicenter collection of preoperative brain MRI acquired during routine clinical care between 2010 and 2022 at six diagnostic centers in Nigeria using 1.5T scanners. Each case includes T1, T1ce, T2, and FLAIR sequences. The cohort provides a relevant stress test because it captures variation in scanner manufacturer and acquisition protocol from a resource-constrained clinical setting; despite image-quality curation, the retained scans exhibit lower signal-to-noise ratio and a higher proportion of artifact-affected voxels than the Western BraTS comparison cohort reported by the dataset authors. This residual heterogeneity makes BraTS-Africa well suited for evaluating how additional controlled corruptions affect pretrained representations.


For each of the four MRI sequences, every clean volume is transformed independently using seven artifact types at five predefined corruption settings. The resulting benchmark therefore contains $95 \times 4 \times 7 \times 5 = 13,300$ corrupted image volumes, in addition to $95 \times 4 = 380$ clean volumes. The benchmark contains 3,325 corrupted volumes per MRI sequence. Representative examples of the seven transforms are shown in \textbf{Appendix}.

\section{Results and Discussion} \label{sec:results}

\subsection{Artifact Sensitivity of Pretrained 3D Encoders}

\begin{figure}[b!]
\centering
\includegraphics[width=0.875\linewidth]{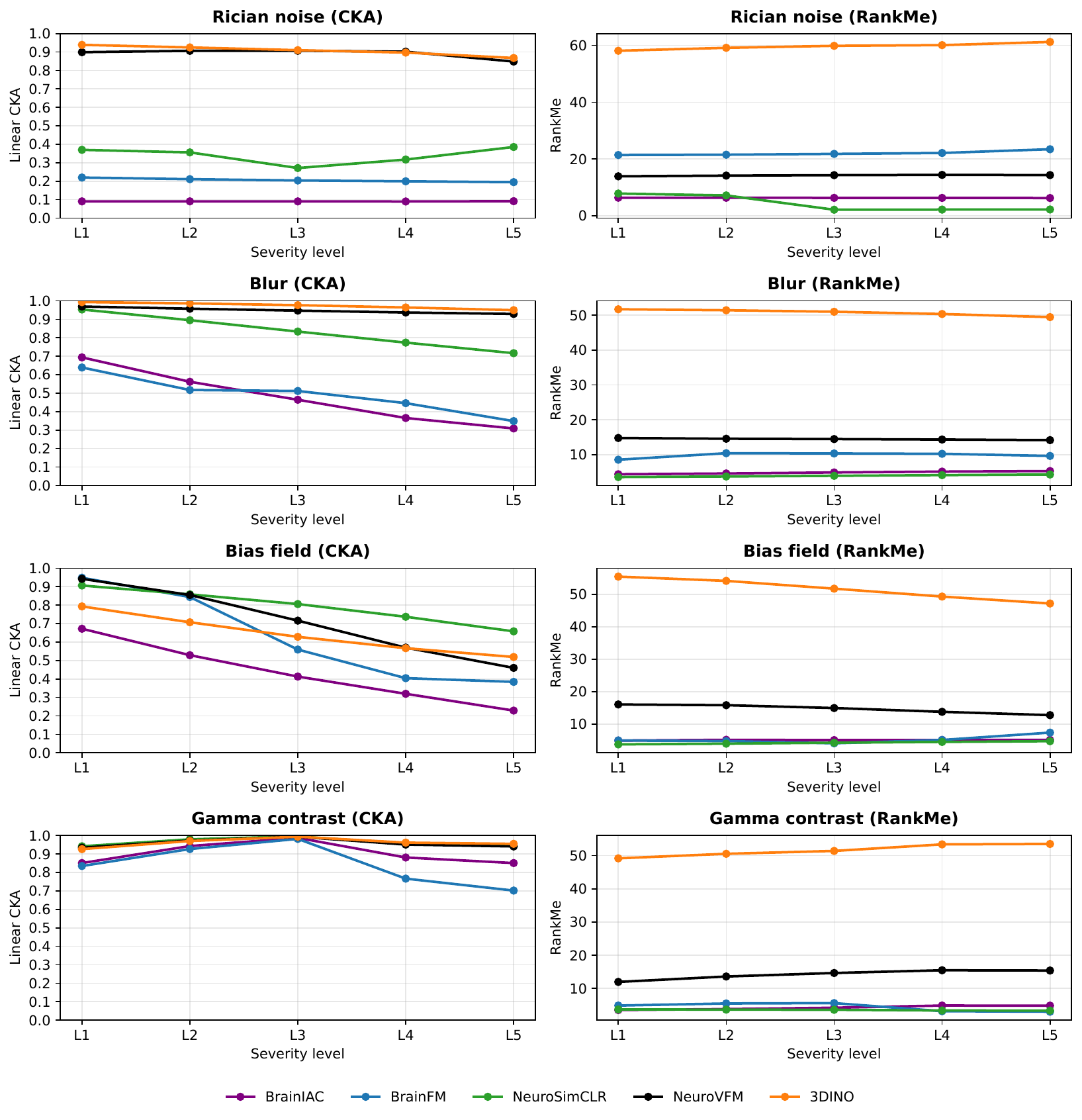}
\vspace{-0.2cm}
\caption{Representation robustness to image-domain artifacts on T1 MRI. Linear CKA and RankMe are shown across five corruption settings for each encoder.}
\label{fig:t1_intensity}
\end{figure}

\begin{figure}[t]
\centering
\includegraphics[width=0.875\linewidth]{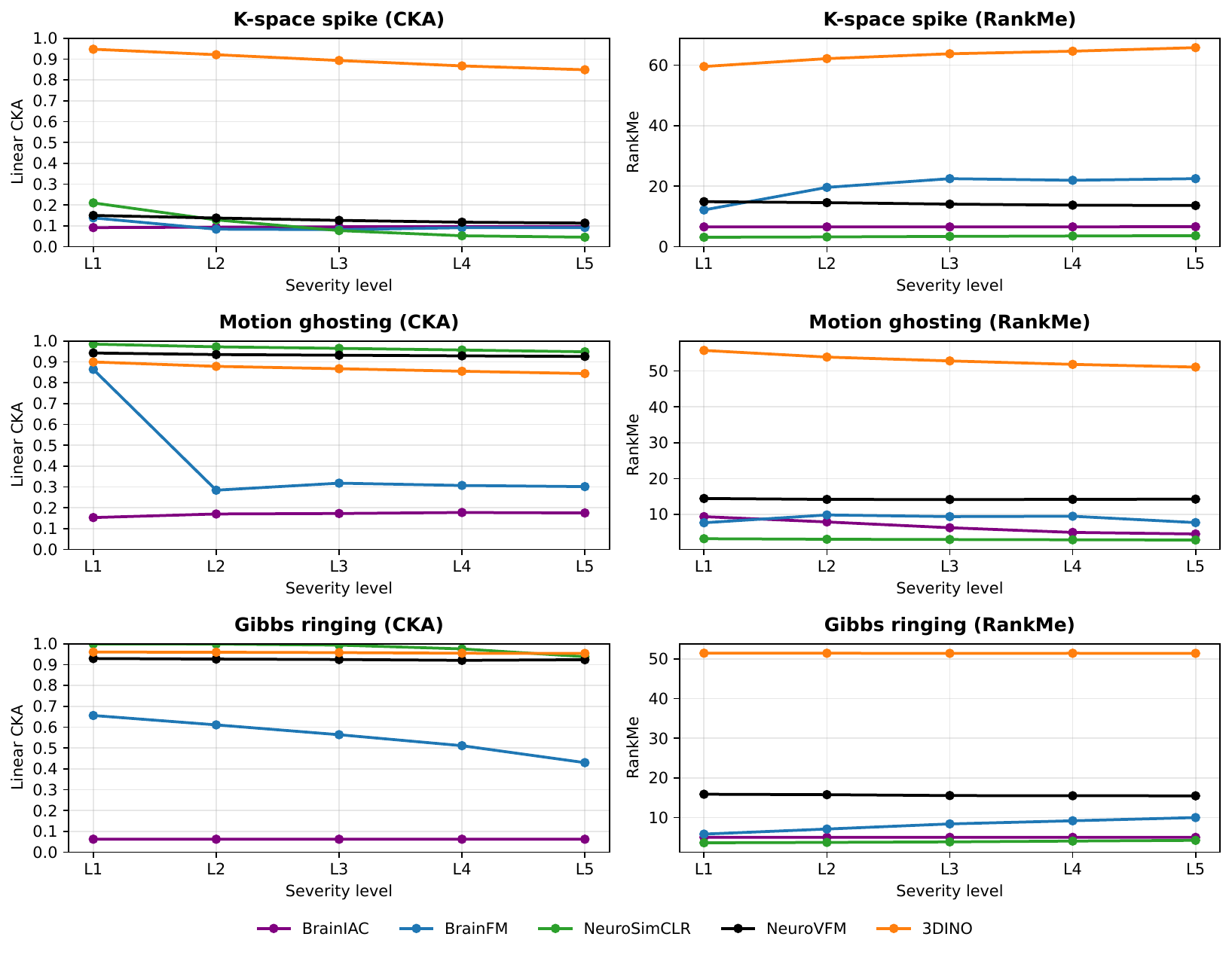}
\vspace{-0.2cm}
\caption{Representation robustness to frequency-domain artifacts on T1 MRI. Linear CKA and RankMe are shown across five corruption settings for each encoder.}
\label{fig:t1_kspace}
\end{figure}

\Cref{fig:t1_intensity,fig:t1_kspace} summarize the representation robustness of the five encoders on T1 MRI. Results for T1ce, T2, and FLAIR are provided in \textbf{Appendix} and show broadly consistent model- and artifact-specific trends.

Across the evaluated corruptions, 3DINO exhibits the most consistently stable representations. Its linear CKA remains high under Rician noise, Gaussian blur, gamma transformation, periodic ghosting, and Gibbs ringing, with a more gradual reduction under k-space spikes and strong bias fields. NeuroVFM also shows high robustness to several structured corruptions, particularly blur, gamma transformation, ghosting, and Gibbs ringing, but is markedly more sensitive to k-space spikes. These results suggest that large-scale pretraining and broad anatomical or modality coverage may improve robustness, although architecture, preprocessing, training objective, and dataset scale differ simultaneously and their individual effects cannot be isolated from this comparison.

BrainIAC is consistently the most artifact-sensitive encoder. Its CKA is low for most corruptions and approaches zero under frequency-domain perturbations. BrainFM exhibits intermediate but strongly artifact-dependent robustness: it remains comparatively stable at mild corruption settings but degrades progressively under blur, bias field, ghosting, and Gibbs truncation. Neuro-SimCLR is comparatively robust to periodic ghosting and Gibbs ringing, but is more sensitive to Rician noise, bias-field variation, and k-space spikes. The relative ordering is largely preserved across modalities, although BrainFM and Neuro-SimCLR display greater modality dependence than 3DINO and NeuroVFM.

The artifact type is as important as the nominal corruption level. K-space spikes produce substantial representation changes for all encoders except 3DINO, whereas periodic ghosting and Gibbs ringing sharply separate the models: 3DINO, NeuroVFM, and Neuro-SimCLR remain comparatively stable, while BrainFM and especially BrainIAC deteriorate strongly. Among the image-domain corruptions, bias field produces the most consistent severity-dependent decline across models. Rician noise strongly affects BrainIAC and BrainFM but has a smaller effect on 3DINO and NeuroVFM. Gaussian blur causes a gradual loss of similarity, particularly for BrainIAC and BrainFM.

The gamma results should be interpreted differently from the other curves. Because the evaluated parameters include values below and above the identity transformation $\gamma=1$, the five settings do not form a single monotonic severity axis. CKA is generally highest for the conditions closest to the identity transformation and decreases for stronger brightening or darkening, indicating sensitivity to the magnitude of the contrast change rather than to the level index itself.

\subsection{Geometric Drift Without Consistent Dimensional Collapse}

CKA and RankMe reveal complementary aspects of representation robustness. Across many artifact--model combinations, CKA decreases substantially with corruption strength while RankMe remains nearly constant. This is clear for blur, bias field, k-space spikes, and ghosting, where the relational geometry of the representation space changes without a corresponding reduction in rank.

This divergence indicates that the dominant effect of MRI artifacts is not necessarily a collapse onto a lower-dimensional subspace. Instead, the artifacts reorganize the relative positions of subjects within the embedding space while preserving a similar spread of singular values. RankMe alone would fail to identify several prominent robustness failures detected by CKA. Conversely, modest RankMe changes should not be interpreted as information loss, since it measures spectral dispersion rather than the clinical relevance of the retained directions.

Absolute RankMe values differ substantially across models, with 3DINO showing the highest values and BrainIAC and Neuro-SimCLR generally showing lower values. These differences should not be interpreted as a ranking of representation quality, because RankMe is affected by feature dimensionality, sample size, and the baseline spectrum of each encoder. The more informative observation is the within-model trajectory across corruption settings. Most models exhibit only modest changes, further supporting the conclusion that artifact sensitivity is expressed primarily as geometric drift rather than dimensional collapse.

\begin{figure}[t]
\centering
\includegraphics[width=0.75\linewidth,height=0.49\linewidth]{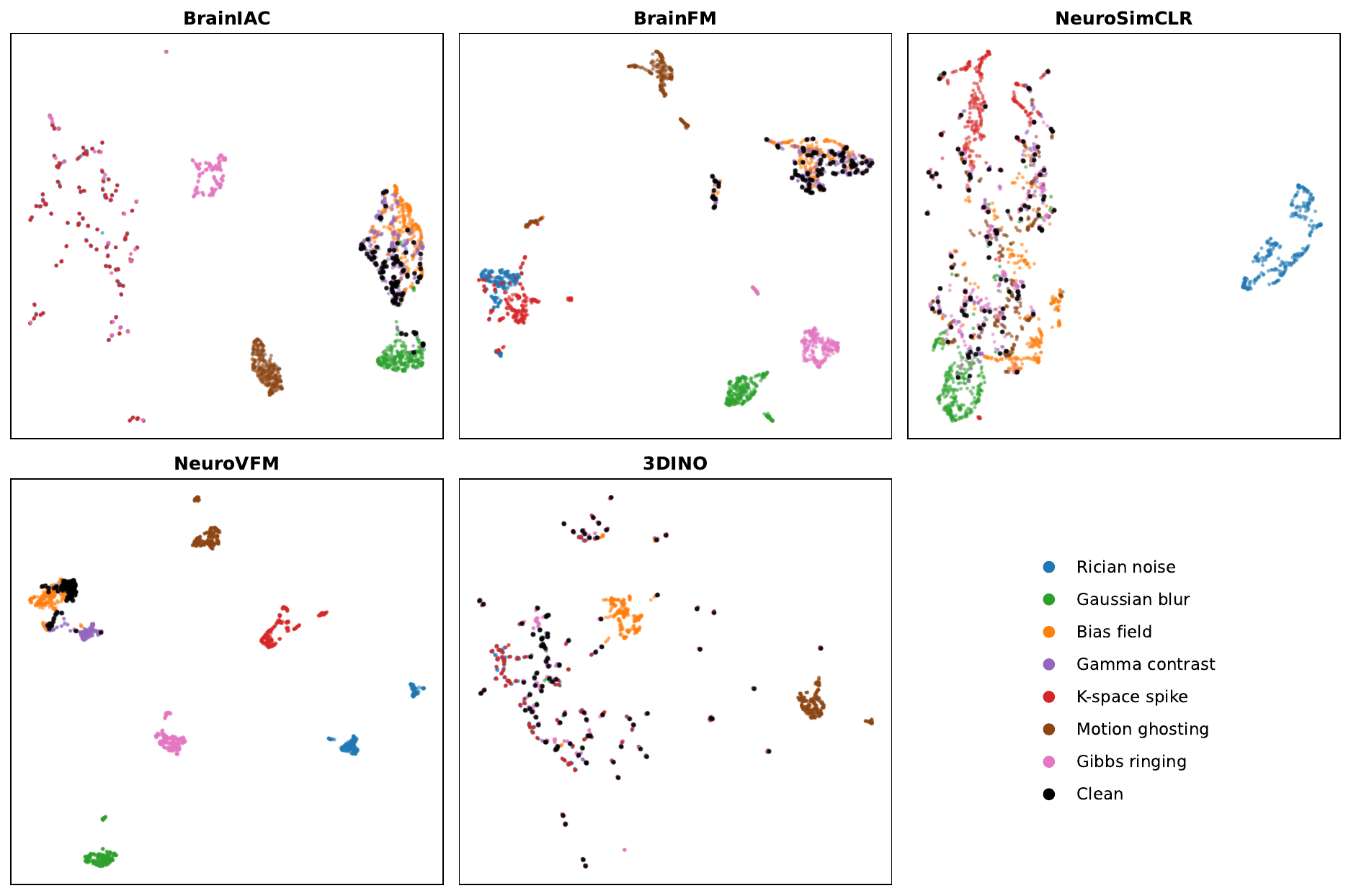}
\vspace{-0.15cm}
\caption{UMAP projections of T1 representations from five encoders under seven artifacts. Clean and artifact-corrupted samples are embedded jointly for each encoder.}
\label{fig:umap_t1}
\end{figure}

The T1 UMAP projections in \Cref{fig:umap_t1} support this interpretation. BrainFM and NeuroVFM form distinct artifact-associated groups, indicating that their embeddings encode corruption-specific variation. BrainIAC shows strong separation for spike and Gibbs-corrupted samples, consistent with its low CKA. In contrast, 3DINO displays greater overlap between clean and corrupted samples, in agreement with its generally higher representational similarity.

The UMAP and CKA results are not expected to agree perfectly. Linear CKA is computed after centering and is invariant to global translation, isotropic scaling, and orthogonal changes of basis, whereas UMAP may visibly separate groups undergoing global distribution shifts. An encoder can therefore retain high CKA while still exhibiting artifact-dependent offsets in a nonlinear two-dimensional projection. The UMAPs are consequently used as qualitative support rather than as an independent quantitative robustness measure.

\subsection{Relation to Segmentation Stability}

\begin{figure}[t]
\centering
\includegraphics[width=0.5\linewidth]{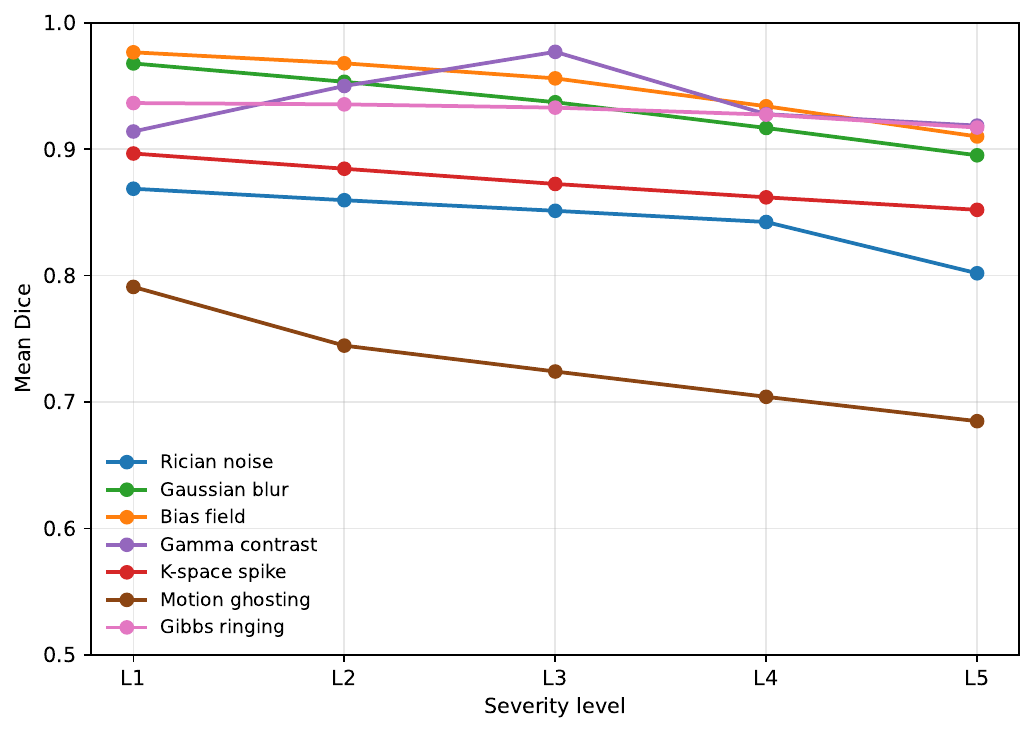}
\vspace{-0.2cm}
\caption{Segmentation consistency under artifact corruption. Dice similarity between clean and corrupted TumorSynth predictions is averaged across labels 1--18 and cases.}
\label{fig:seg_consistency}
\end{figure}

\Cref{fig:seg_consistency} reports the Dice similarity between TumorSynth predictions from clean and corrupted inputs. Since TumorSynth is independent of the five evaluated encoders, these results do not constitute downstream validation of the model representations. Instead, they provide a task-level reference for the relative effects of the artifacts reflected in an automated segmentation pipeline.

Periodic ghosting produces the largest reduction in prediction consistency, with mean Dice decreasing steadily from $0.79$ at L1 to $0.69$ at L5. Rician noise causes the second-largest degradation, followed by k-space spikes. Blur, bias field, Gibbs ringing, and gamma transforms retain comparatively high consistency over the evaluated settings with some degradation at stronger corruptions.

The strong effect of ghosting is also visible in the representative segmentation maps provided in \textbf{Appendix}, where ghost replicas lead to spatially implausible predicted regions outside the main anatomy. The remaining corruptions primarily induce smaller boundary and regional changes.

The task-level ranking only partially matches the representation-level findings. Ghosting strongly affects the segmentation model and also produces large representation changes for BrainFM and BrainIAC, but it has little effect on 3DINO, NeuroVFM, and Neuro-SimCLR. Conversely, k-space spikes strongly perturb several encoder representations while producing a moderate segmentation consistency loss. These differences show that artifact sensitivity is model- and task-dependent: an artifact may alter a pretrained representation without producing the same degree of instability in a separate segmentation pipeline.

\section{Conclusion} \label{sec:conclusion}

We presented a controlled study of how MRI artifacts affect the representations learned by five pretrained 3D medical-image encoders. Robustness varied substantially across models and corruption types: 3DINO was the most consistently stable, whereas other encoders showed distinct artifact-specific sensitivities. Across many conditions, CKA declined while RankMe remained comparatively stable, indicating that artifacts often distort representation geometry without causing dimensional collapse.

The segmentation analysis further showed that task-level sensitivity only partially follows representation-level robustness. These findings demonstrate that neither pretraining scale nor domain specialization alone guarantees robustness to MRI artifacts, and highlight the need to assess representation stability before deploying foundation models across heterogeneous clinical imaging conditions.

\section*{Acknowledgements} 
This project is supported by the Pioneer Centre for AI, funded by the Danish National Research Foundation (grant number P1).

\bibliographystyle{splncs}
\bibliography{references}

\newpage
\appendix
\chapter*{Appendix} \label{sec:appendix}

\section{Utilized Encoders}

\subsection{3DINO}

3DINO adapts the DINO self-distillation framework to volumetric medical images using complementary image-level and masked patch-level objectives. Its 3D ViT-L/16 encoder was pretrained on approximately 100,000 unlabeled volumes from 35 studies, comprising MRI, CT, and brain PET images from more than 10 anatomical regions. It therefore represents a broadly pretrained medical encoder rather than a brain- or MRI-specific model. Following the released inference pipeline, each volume is resized to $112^3$ voxels and intensities are linearly rescaled to $[-1,1]$. We use the 1024-dimensional class token from the final transformer layer, that is $\phi_{\mathrm{3DINO}}(\mathcal{I}) \in \mathbb{R}^{1024}$.

\subsection{BrainIAC}

BrainIAC is a brain-MRI-specific foundation model based on a 3D ViT-B/16 encoder pretrained with the SimCLR objective. The model was pretrained on 32,015 multiparametric brain MRI scans selected from 16 datasets spanning 10 neurological conditions. The larger collection of 48,965 scans reported in the study includes data used across model development and downstream evaluation and does not correspond solely to the self-supervised pretraining set. Following the released inference pipeline, volumes are resized to $96^3$ voxels and $z$-score normalized using the non-zero voxels. We extract the 768-dimensional class token from the final transformer layer, so $\phi_{\mathrm{BrainIAC}}(\mathcal{I})\in\mathbb{R}^{768}$.

\subsection{BrainFM}

BrainFM differs from the other evaluated models by using a 3D U-Net-like, fully convolutional architecture. It is trained jointly for image synthesis, anatomical segmentation, cortical-distance estimation, bias-field estimation, and atlas registration. Its training strategy combines on-the-fly intra-subject synthesis with real--synthetic mixing to expose the model to variations in modality, contrast, resolution, orientation, deformation, and image quality. In Setup I, the 11 source datasets contain 8,675 modality-specific images before the dataset-wise 90/10 training--test split, corresponding to approximately 7,800 real training images. This count treats different modalities from the same subject as separate images and excludes the unlimited synthetic samples generated during training.

Following the published preprocessing protocol, input images are skull-stripped and resampled to $1$-mm isotropic resolution. BrainFM is fully convolutional and consequently does not require a single fixed matrix size; the released implementation supports whole-volume and tiled inference. Input volumes are min--max normalized over the full volume range. Since BrainFM is not originally defined as a global image encoder, we extract the released encoder feature map and average it over its spatial dimensions. The resulting representation for the bottleneck feature map (deepest encoder layer) is therefore $\phi_{\mathrm{BrainFM}}(\mathcal{I})\in\mathbb{R}^{2048}$. The global pooling operation is specific to our representation analysis and is not part of BrainFM's original task-specific inference procedure.

\subsection{NeuroVFM}

NeuroVFM is a generalist neuroimaging encoder pretrained on 5.24 million clinical MRI and CT volumes from 566,915 studies acquired over more than two decades of routine clinical care. It uses a 3D ViT-B encoder with $4\times16\times16$ volumetric patches and is trained using volumetric JEPA. Vol-JEPA learns to predict latent representations of masked target regions from visible context regions. The preprocessing pipeline quantizes the clinical volumes before tokenization. Because the model was designed for heterogeneous clinical series, it does not require all input volumes to be resized to one fixed 3D matrix. We use the official preprocessing object associated with the released checkpoint and obtain a volume-level representation by averaging the final-layer patch tokens as $\phi_{\mathrm{NeuroVFM}}(\mathcal{I})\in\mathbb{R}^{768}$.

\subsection{Neuro-SimCLR}

Neuro-SimCLR uses a 3D ResNet-18 encoder pretrained with SimCLR on 44,958 structural brain MRI scans from 18,759 patients across 11 public datasets and multiple neurological conditions. It complements BrainIAC by combining an instance-contrastive objective with a convolutional rather than transformer-based backbone. We follow the released TurboPrep-based preprocessing and inference pipeline. The image axes are reordered to the orientation expected by the encoder, the brain volume is cropped using the coordinates prescribed by the released implementation, and intensities are $z$-score normalized over foreground voxels. No additional isotropic resizing is introduced in our experiments. The 512-dimensional output of the encoder's global average-pooling layer is used as the representation to obtain $\phi_{\mathrm{NeuroSimCLR}}(\mathcal{I})\in\mathbb{R}^{512}$.

\section{Additional Results}

\begin{figure}[t]
\centering
\includegraphics[width=0.875\linewidth]{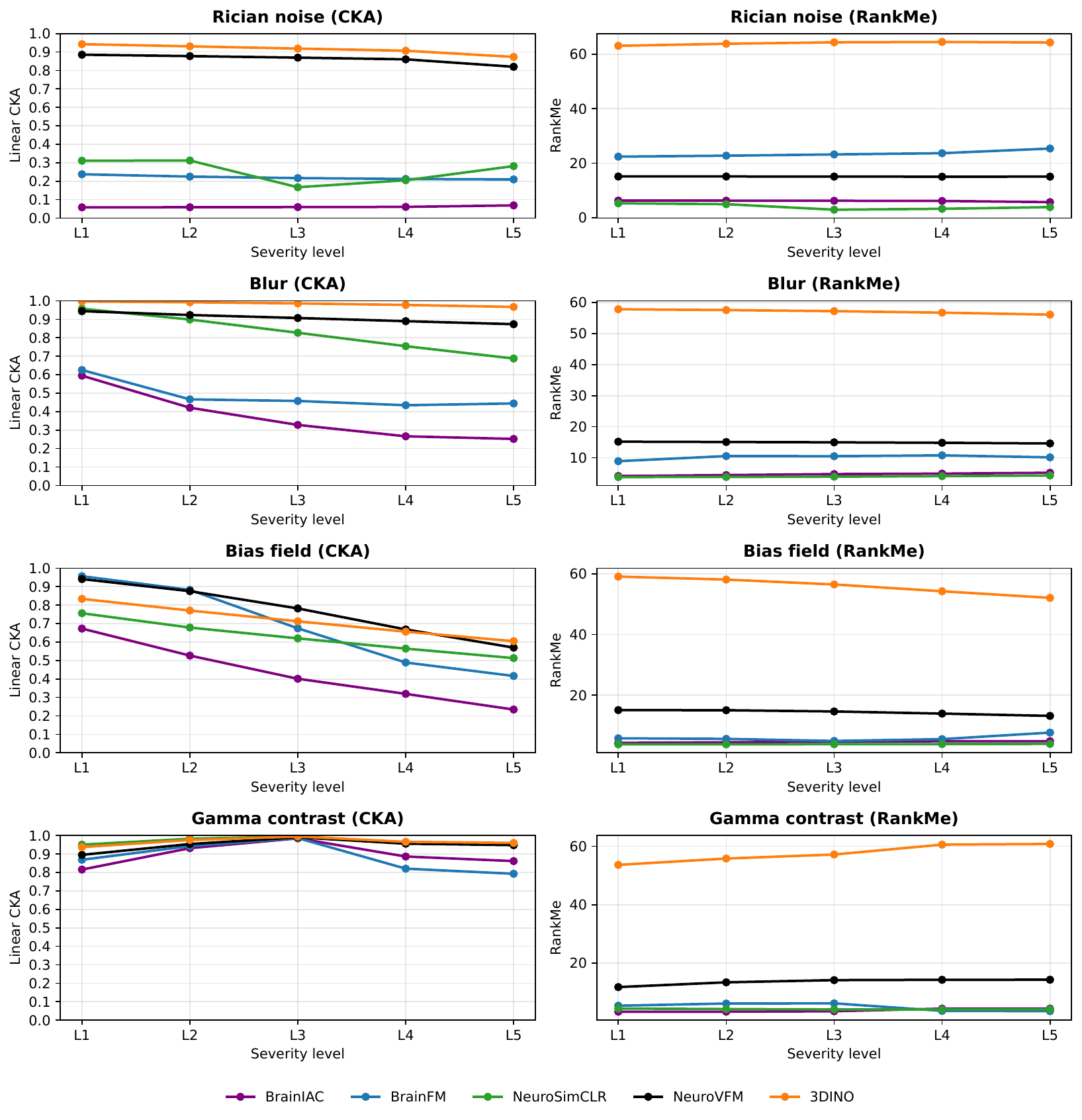}
\vspace{-0.2cm}
\caption{Representation robustness to image-domain artifacts on T1ce MRI. Linear CKA and RankMe are shown across five corruption settings for each encoder.}
\label{fig:t1ce_intensity}
\end{figure}

\begin{figure}[t]
\centering
\includegraphics[width=0.875\linewidth]{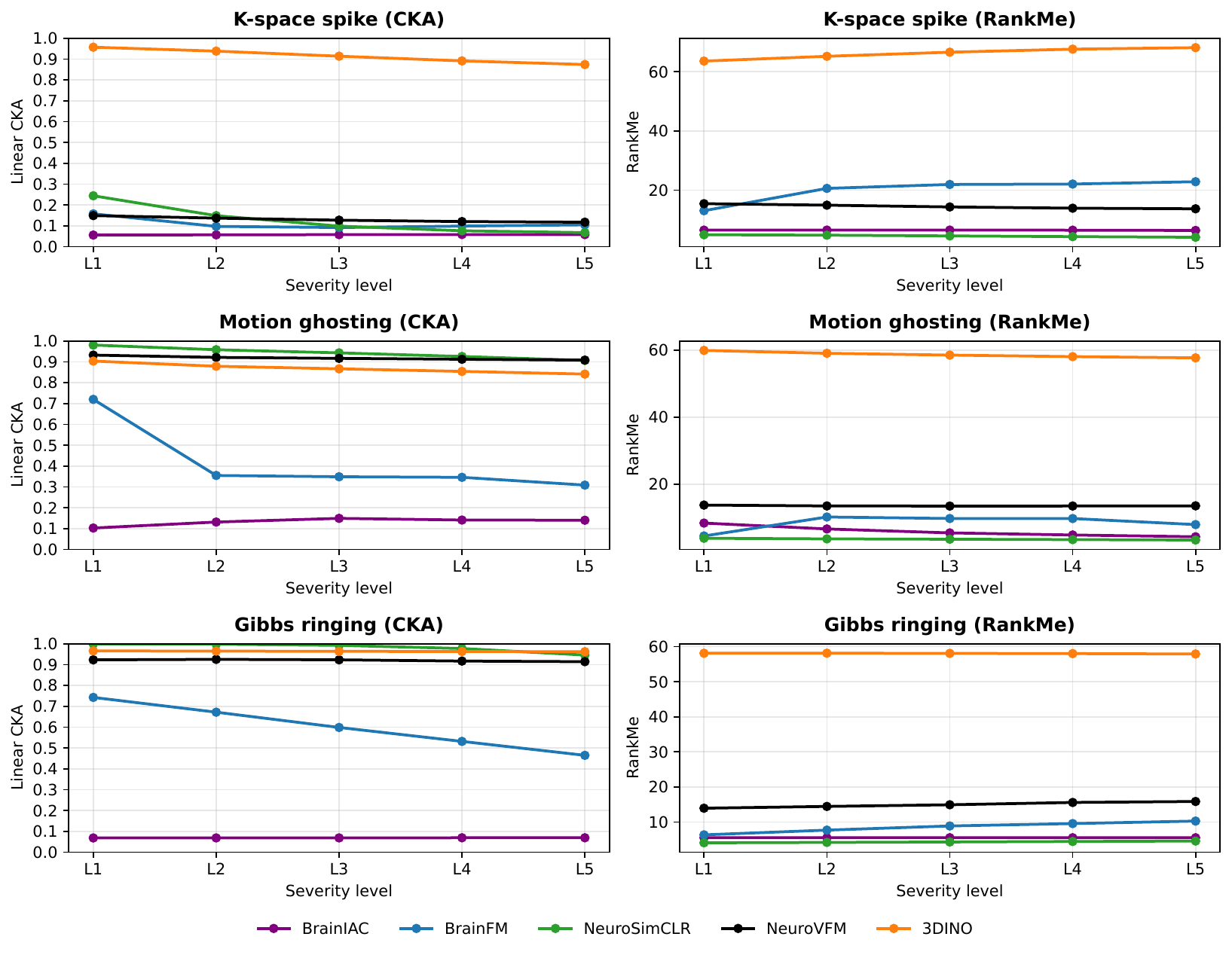}
\vspace{-0.2cm}
\caption{Representation robustness to frequency-domain artifacts on T1ce MRI. Linear CKA and RankMe are shown across five corruption settings for each encoder.}
\label{fig:t1ce_kspace}
\end{figure}

\begin{figure}[t]
\centering
\includegraphics[width=0.875\linewidth]{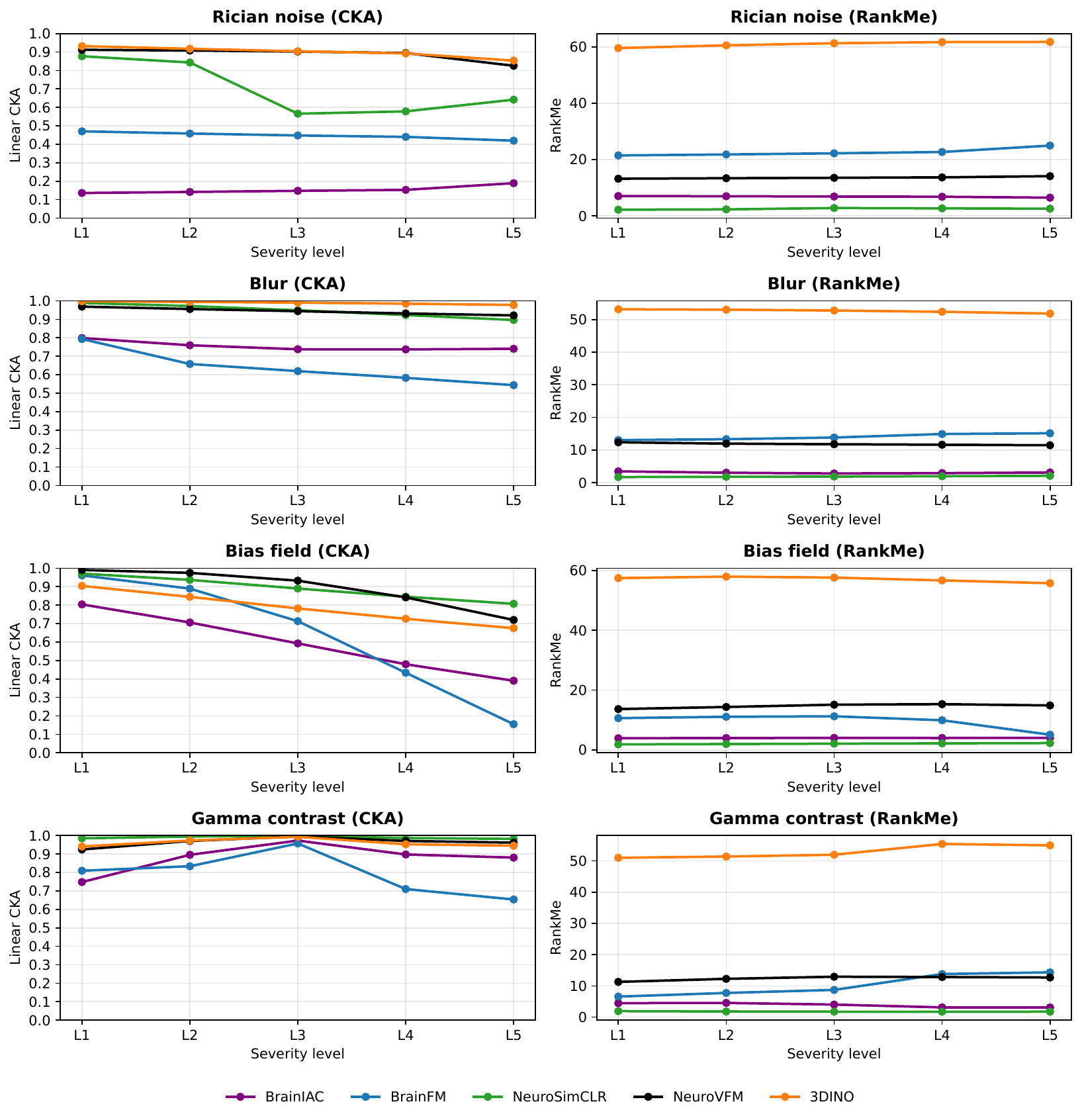}
\vspace{-0.2cm}
\caption{Representation robustness to image-domain artifacts on T2 MRI. Linear CKA and RankMe are shown across five corruption settings for each encoder.}
\label{fig:t2_intensity}
\end{figure}

\begin{figure}[t]
\centering
\includegraphics[width=0.875\linewidth]{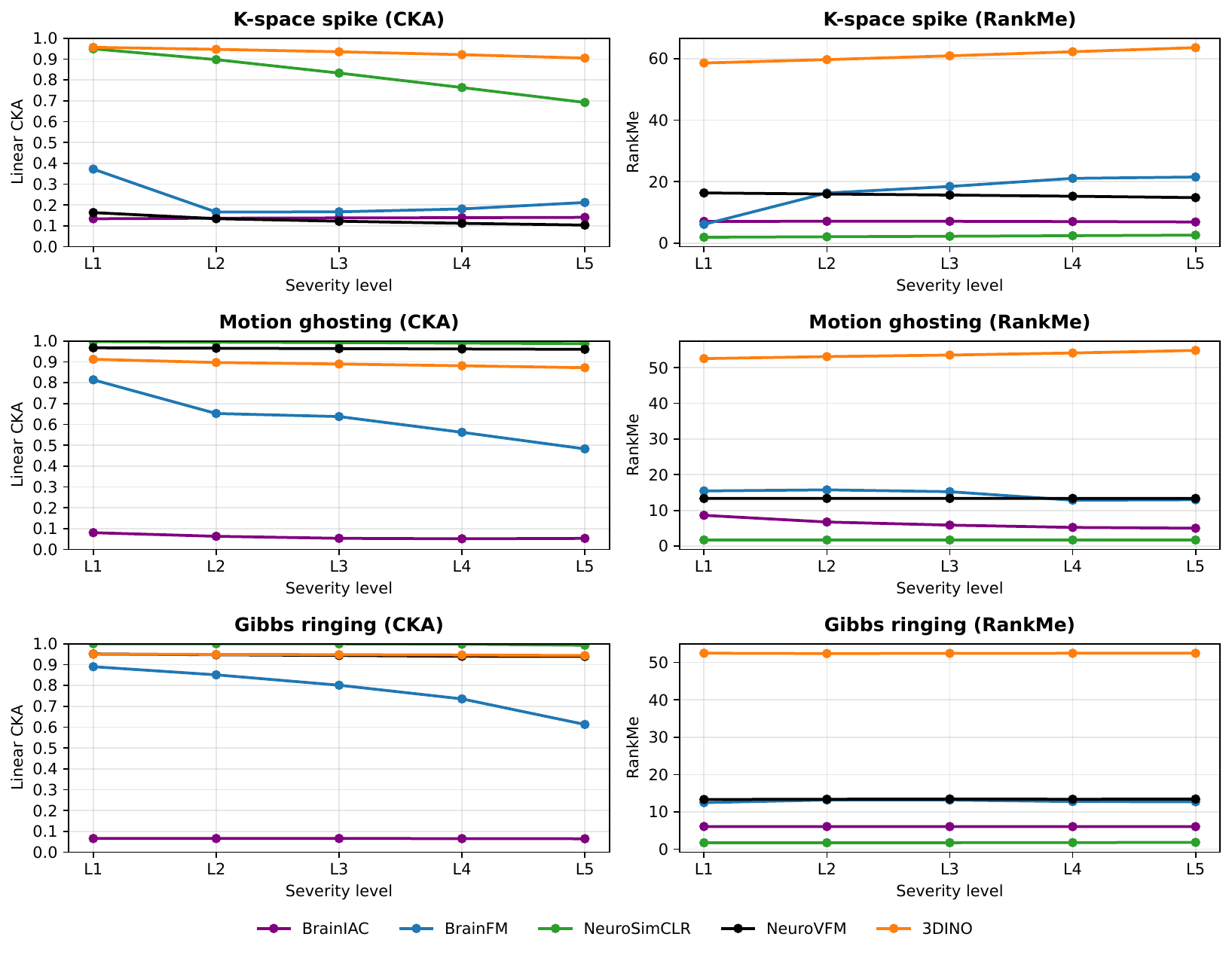}
\vspace{-0.2cm}
\caption{Representation robustness to frequency-domain artifacts on T2 MRI. Linear CKA and RankMe are shown across five corruption settings for each encoder.}
\label{fig:t2_kspace}
\end{figure}

\begin{figure}[t]
\centering
\includegraphics[width=0.875\linewidth]{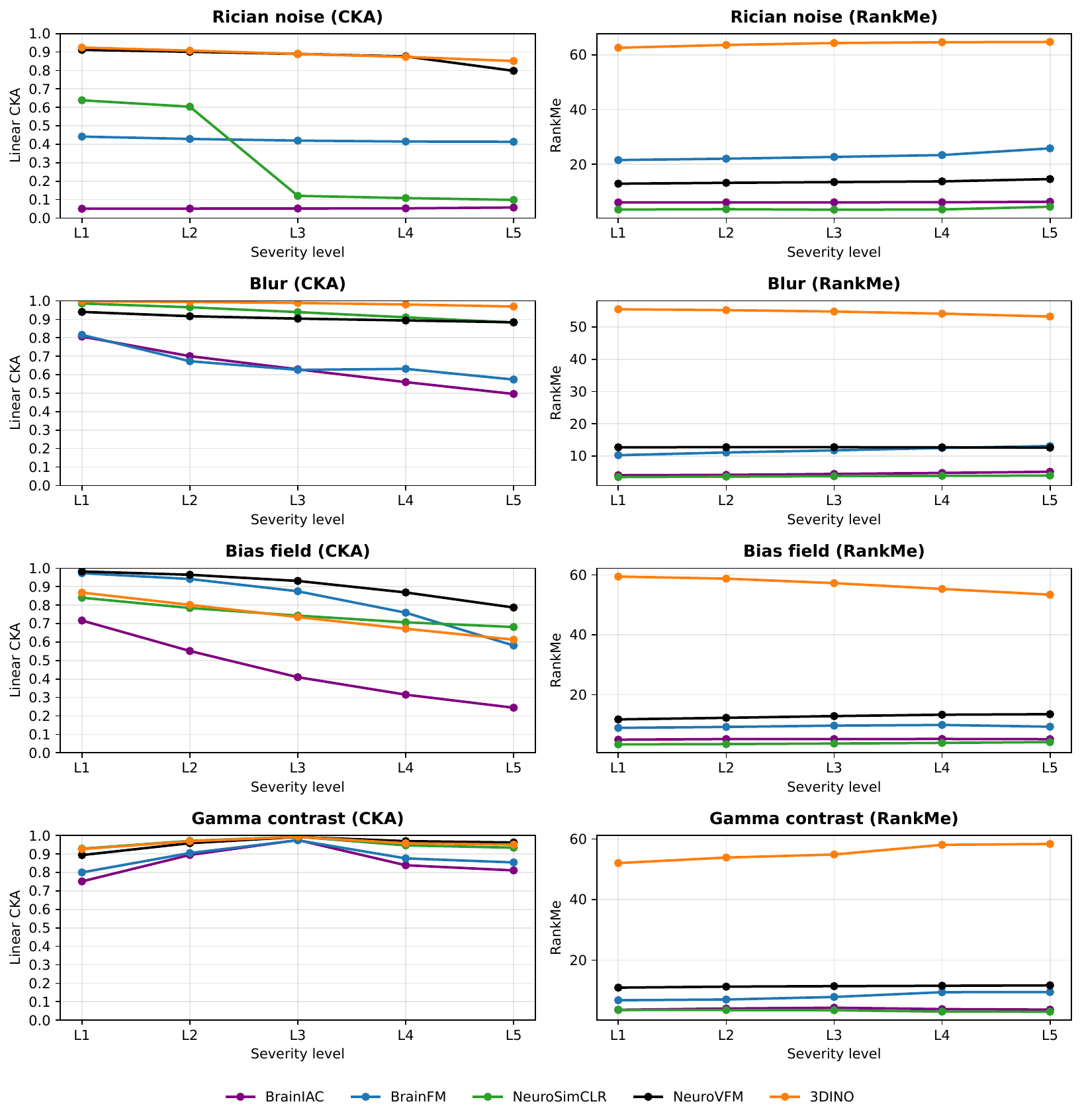}
\vspace{-0.2cm}
\caption{Representation robustness to image-domain artifacts on FLAIR MRI. Linear CKA and RankMe are shown across five corruption settings for each encoder.}
\label{fig:flair_intensity}
\end{figure}

\begin{figure}[t]
\centering
\includegraphics[width=0.875\linewidth]{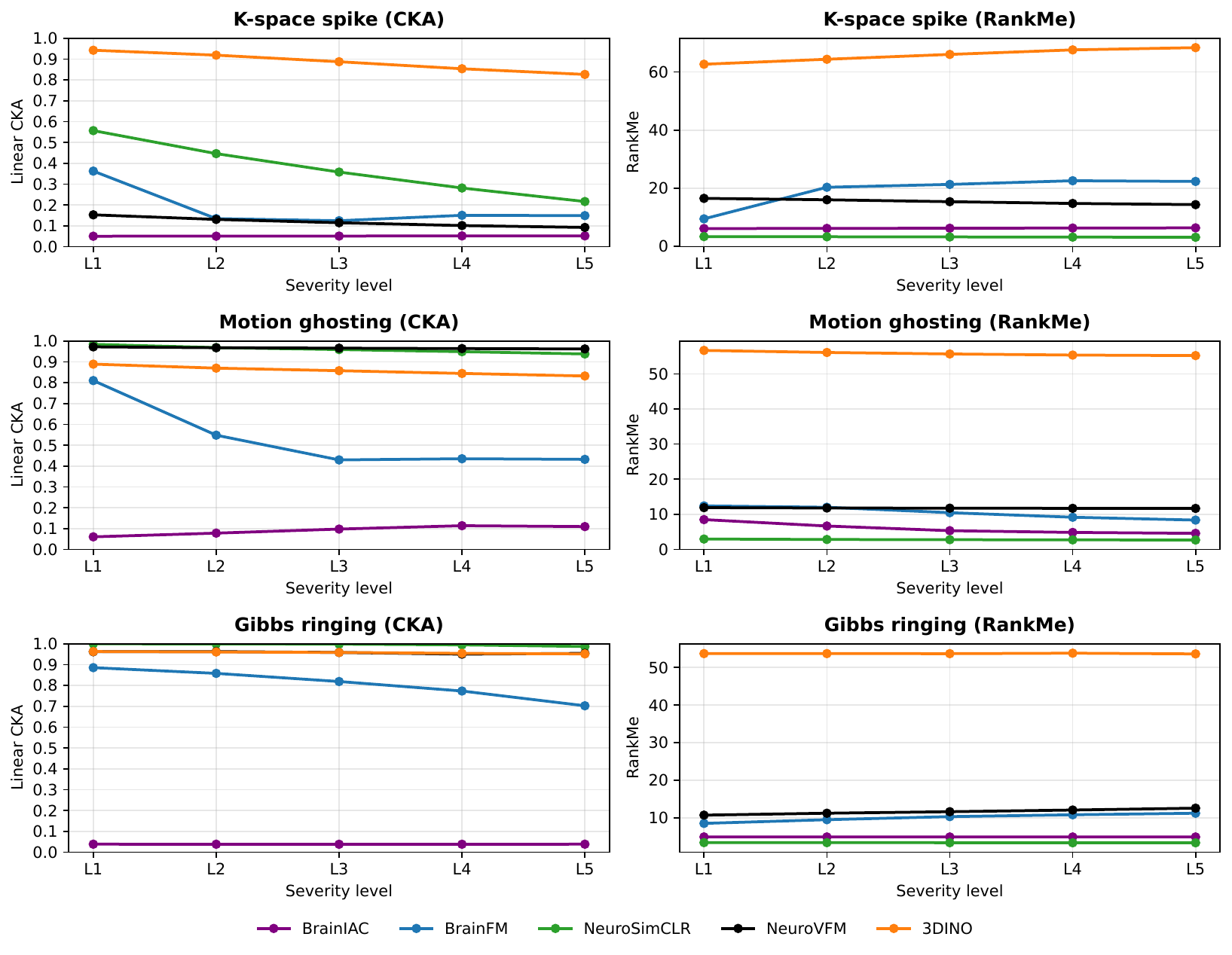}
\vspace{-0.2cm}
\caption{Representation robustness to frequency-domain artifacts on FLAIR MRI. Linear CKA and RankMe are shown across five corruption settings for each encoder.}
\label{fig:flair_kspace}
\end{figure}

\begin{figure}[t]
\centering
\includegraphics[width=0.75\linewidth,height=0.49\linewidth]{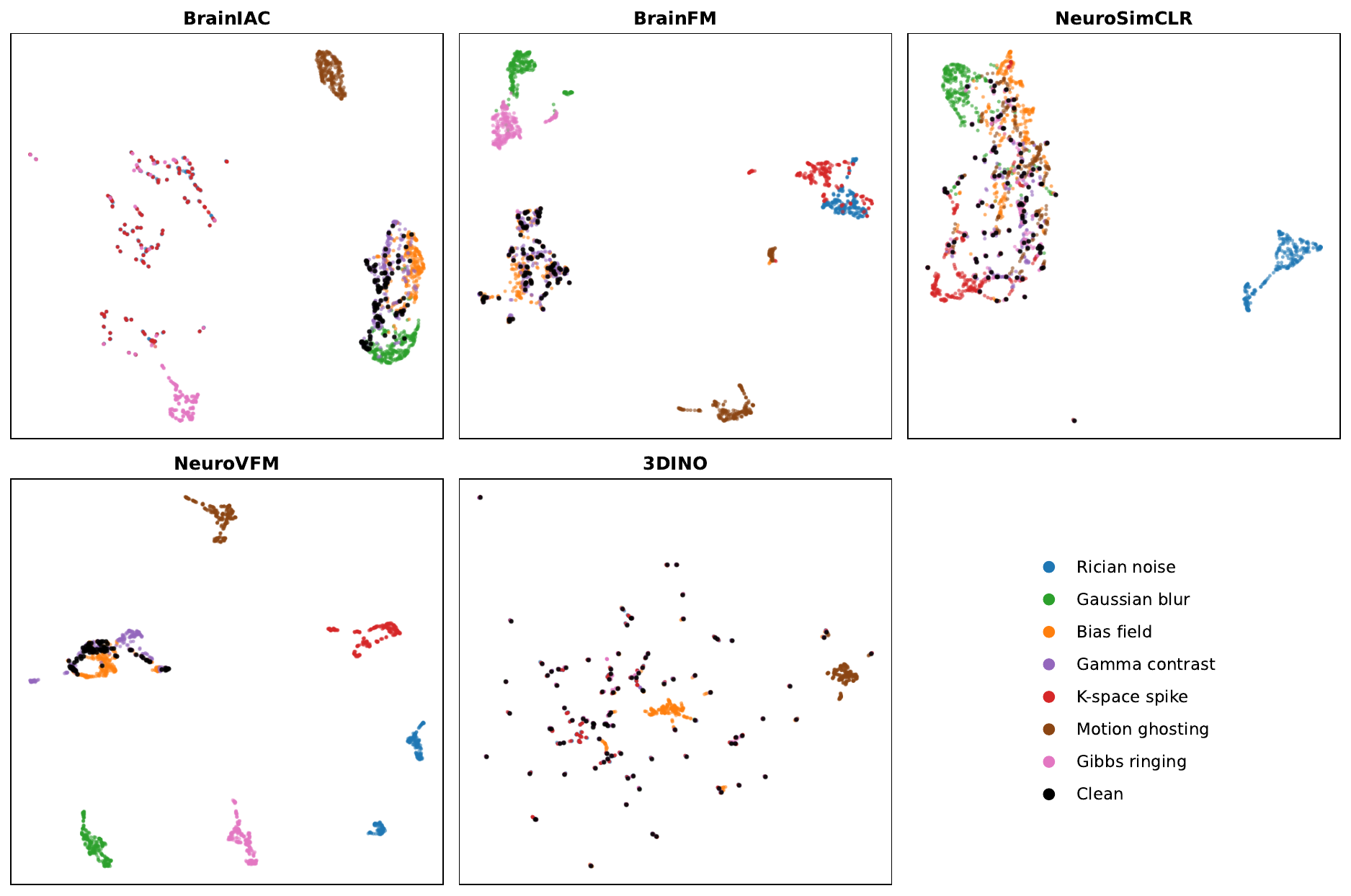}
\vspace{-0.15cm}
\caption{UMAP projections of T1ce representations from five encoders under seven artifacts. Clean and artifact-corrupted samples are embedded jointly for each encoder.}
\label{fig:umap_t1ce}
\end{figure}

\begin{figure}[t]
\centering
\includegraphics[width=0.75\linewidth,height=0.49\linewidth]{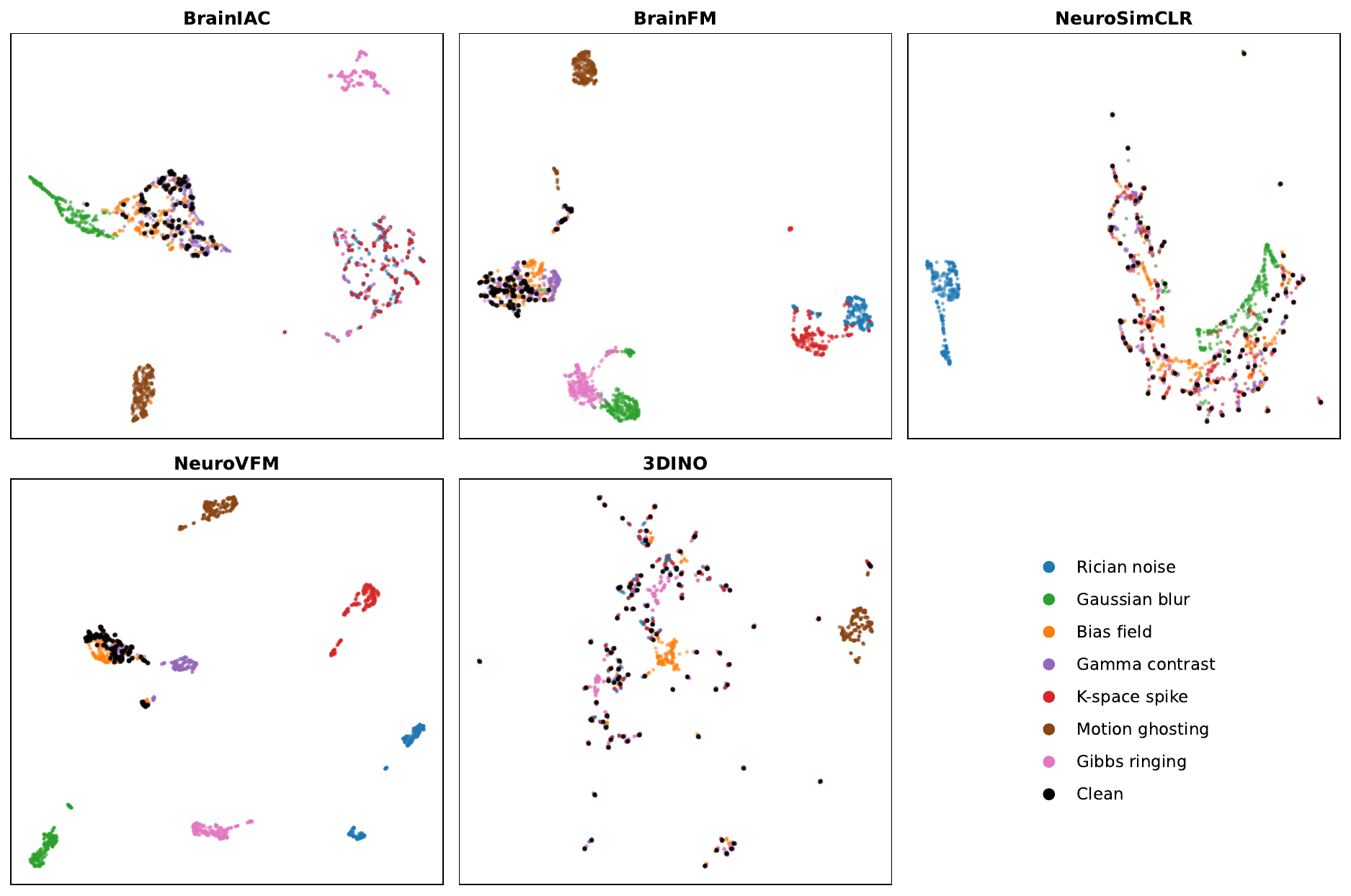}
\vspace{-0.15cm}
\caption{UMAP projections of T2 representations from five encoders under seven artifacts. Clean and artifact-corrupted samples are embedded jointly for each encoder.}
\label{fig:umap_t2}
\end{figure}

\begin{figure}[t]
\centering
\includegraphics[width=0.75\linewidth,height=0.49\linewidth]{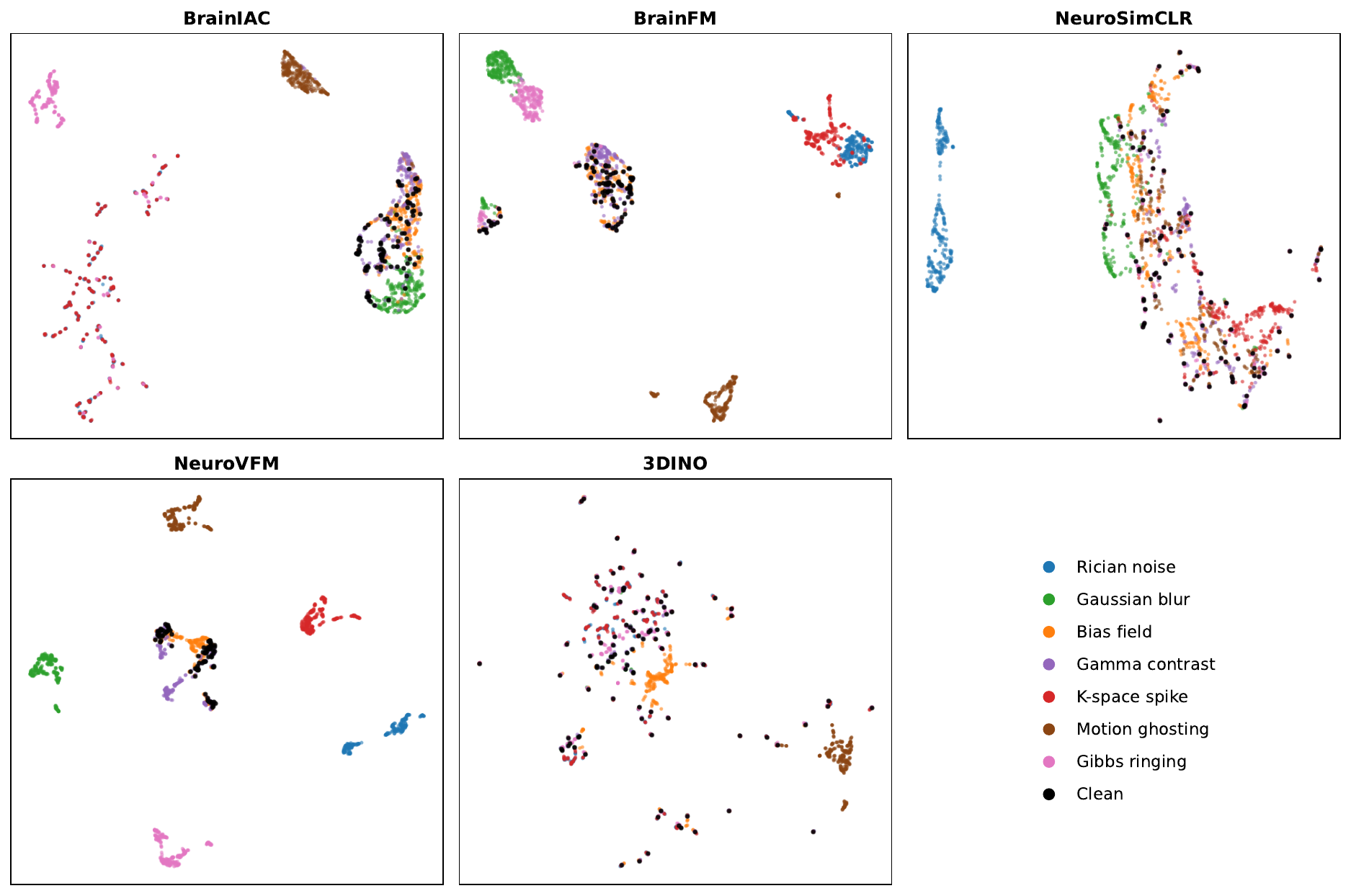}
\vspace{-0.15cm}
\caption{UMAP projections of FLAIR representations from five encoders under seven artifacts. Clean and artifact-corrupted samples are embedded jointly for each encoder.}
\label{fig:umap_flair}
\end{figure}

\begin{figure}[t]
\centering
\includegraphics[width=0.8\linewidth]{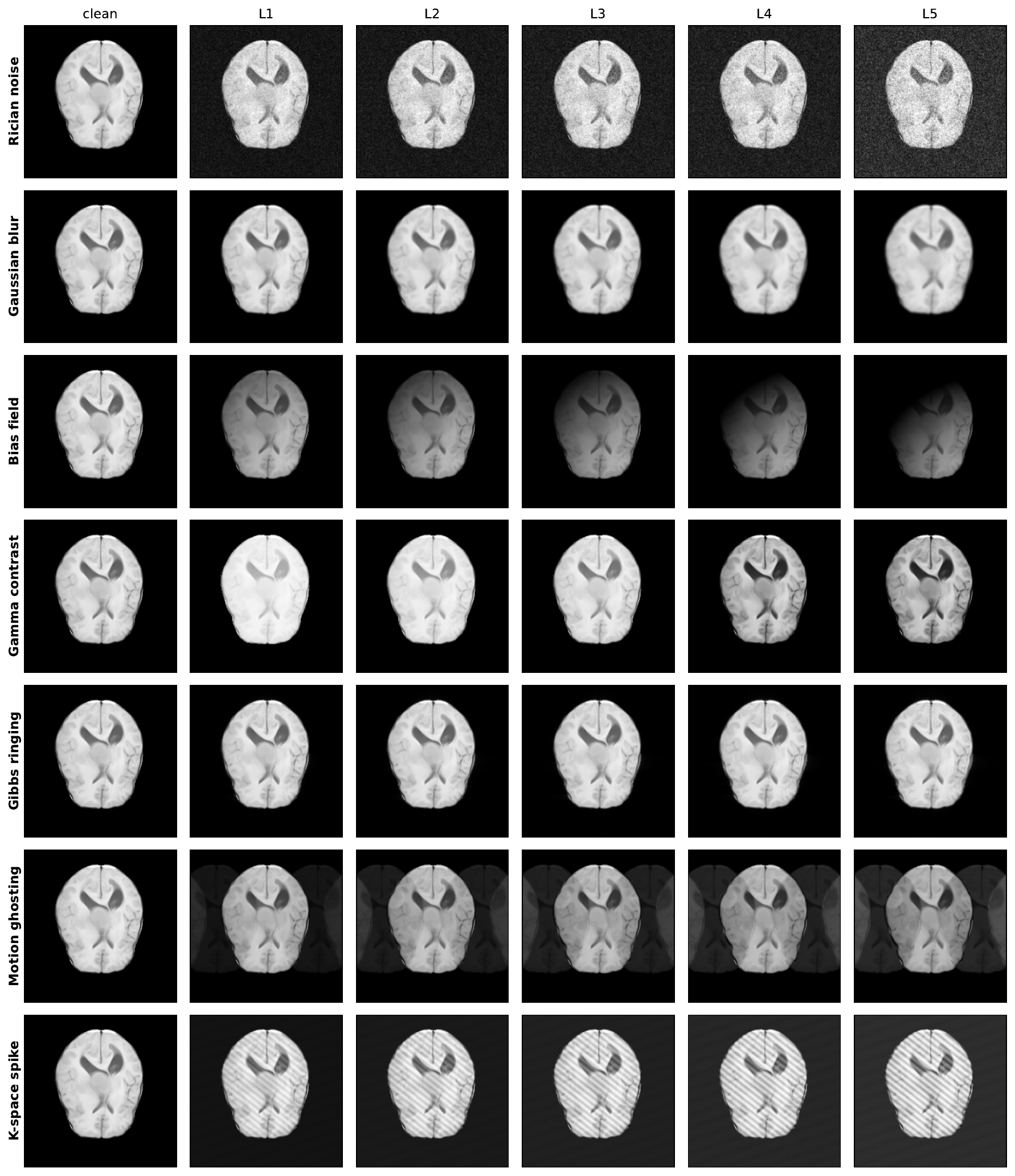}
\vspace{-0.2cm}
\caption{Middle slices of representative T1 MRI under the seven simulated artifacts. Each row corresponds to one artifact type, and columns show the clean image followed by the five predefined corruption settings.}
\label{fig:t1_artifacts}
\end{figure}

\begin{figure}[t]
\centering
\includegraphics[width=0.8\linewidth]{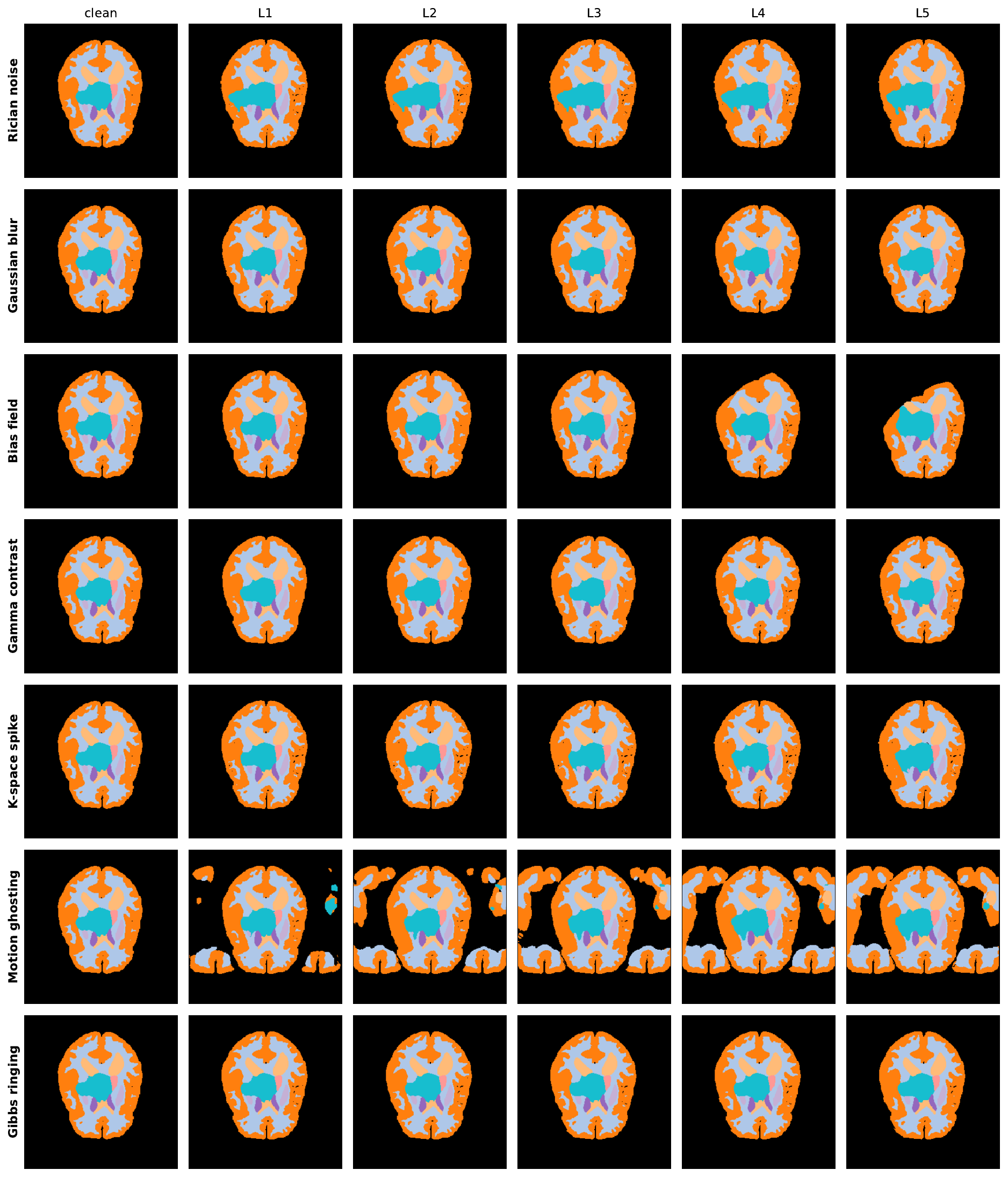}
\vspace{-0.2cm}
\caption{Middle slices of representative TumorSynth predictions under artifact corruption. Each row corresponds to one artifact type, and columns show the 18 predicted anatomical and tumor classes for the clean image and five corruption settings.}
\label{fig:t1_masks}
\end{figure}

\end{document}